\documentclass[letterpaper]{article} 
\usepackage[preprint]{preprint_format}  
\usepackage[hyphens]{url}  
\usepackage{graphicx} 
\usepackage{natbib}  
\usepackage{caption} 
\usepackage{amssymb}
\usepackage{amsmath}

\usepackage{booktabs}
\usepackage{multirow}

\title{PL-SCEA: Reconfiguring Pretrained Attention for Few-Shot Industrial Anomaly Detection}

\author{
   Xiaoyu Yang\textsuperscript{\rm 1},
   Qixing Wu\textsuperscript{\rm 1},
   Huixian Zhao\textsuperscript{\rm 1},
   Changlong Jin\textsuperscript{\rm 1}\corresponding
}
\affiliations{
   \textsuperscript{\rm 1} School of Airspace Science and Engineering, Shandong University, Shandong, China\\
   \{yangxiaoyu,wuqixing,zhaohuixian\}@mail.sdu.edu.cn, cljin@sdu.edu.cn
}

\begin{document}

\maketitle

\begin{abstract}
Vision Foundation Models (VFMs) provide transferable patch representations for few-shot industrial anomaly detection, but their attention computation is typically inherited from pretraining objectives centered on semantic aggregation. This creates a potential mismatch: token relations that support semantic recognition may not adequately expose the localized texture and structural deviations required for anomaly localization. We therefore investigate the hypothesis that the attention computation of a frozen VFM can be reconfigured as a task-relevant component of anomaly detection. We instantiate this idea with Power-Law Self-Correlation Enhanced Attention (PL-SCEA), which retains the semantic context of pretrained query-key attention while constructing token-adaptive self-correlations over contextualized value features. Positive-correlation filtering and power-law reweighting then emphasize relations that are salient relative to each token's relational background, without introducing additional trainable attention projections. The resulting features are modeled by a lightweight variational autoencoder that provides a fixed-size reconstruction-based representation of category-specific normality. The two stages serve complementary roles: attention reconfiguration shapes how local relational deviations are represented, while reconstruction-based modeling converts deviations from learned normality into anomaly scores. Across MVTec AD and VisA, the complete framework achieves competitive image-level detection and consistently strong pixel-level localization across the evaluated few-shot settings. Ablations further show that PL-SCEA improves localization with either the VAE or a memory bank under the tested setting. These results support the view that task-aligned attention reconfiguration can improve the anomaly-localization capability of frozen pretrained representations.
\end{abstract}
\section{Introduction}

Industrial anomaly detection typically trains on normal images to identify defective samples and localize anomalous regions. Feature-matching~\cite{padim,patchcore}, knowledge-distillation~\cite{RD4AD,salehi2021multiresolution}, and reconstruction~\cite{draem} methods perform well on MVTec AD~\cite{mvtec} and VisA~\cite{visa}, but generally require category-specific normal data that cover texture, pose, and structural variation. Rebuilding such datasets for new product categories is costly. Few-shot industrial anomaly detection must therefore establish category-specific normality from only $K$ normal reference images while remaining sensitive to local defects.

In few-shot settings, the limited normal references make the quality and organization of pretrained patch representations particularly important. RADIO~\cite{radio} distills complementary capabilities from DINO~\cite{dino}, CLIP~\cite{clip}, and SAM~\cite{sam}, while RADIOv3~\cite{PHI-S} further improves the spatial detail of dense representations. Nevertheless, these models inherit attention computations optimized primarily for general-purpose semantic representation. Their pretrained features therefore provide a strong starting point, but the way token information is aggregated may remain mismatched with the localized texture and structural deviations that characterize industrial anomalies.

Attention computation determines how each output patch aggregates evidence from the remaining tokens. In the example shown in Figure~\ref{chewinggum_attention_comparison}, the original QK attention preserves semantic context but produces relatively fragmented responses around boundaries and high-contrast regions. Direct QQ, KK, and VV correlations form more spatially continuous patterns, but some also merge locally distinct regions and reduce defect contrast. This observation suggests complementary roles for semantic conditioning and same-type token relations. It further raises a broader question: can the attention computation of a frozen VFM be reconfigured for anomaly localization without discarding the semantic structure learned during pretraining?

Motivated by this question, we argue that the attention computation of a frozen VFM should be treated as a task-relevant design variable rather than an immutable part of the feature extractor. Importantly, modifying attention is not beneficial by default: directly replacing QK attention with QQ, KK, or VV correlations may discard useful semantic conditioning or over-smooth local structures. An anomaly-oriented reconfiguration should instead retain pretrained QK context while selectively reshaping token relations according to their relative relational salience.

We instantiate this design principle with Power-Law Self-Correlation Enhanced Attention (PL-SCEA). PL-SCEA first contextualizes value features through the original QK attention and then computes pairwise self-correlations over the contextualized tokens. Token-wise standardization and power-law reweighting emphasize relations that are salient relative to each token's relational background, and the resulting weights are used to reaggregate the original value tokens. This design preserves pretrained semantic conditioning while altering how local relational evidence is represented, without introducing additional trainable attention projections.

Attention reconfiguration alone does not define how category-specific normality should be represented from a few reference images. We therefore pair PL-SCEA with a lightweight variational autoencoder~\cite{VAE} that reconstructs relation-enhanced normal patch features. The two stages address different parts of the detection pipeline: PL-SCEA shapes how local relational deviations are exposed in the feature space, whereas the reconstruction model summarizes normality and converts deviations from that normality into patch-level anomaly scores. The resulting parametric normality model has a fixed storage footprint and avoids nearest-neighbor retrieval over an explicit feature memory bank.

\begin{figure*}[!htbp]
    \centering
    \includegraphics[width=1\linewidth]{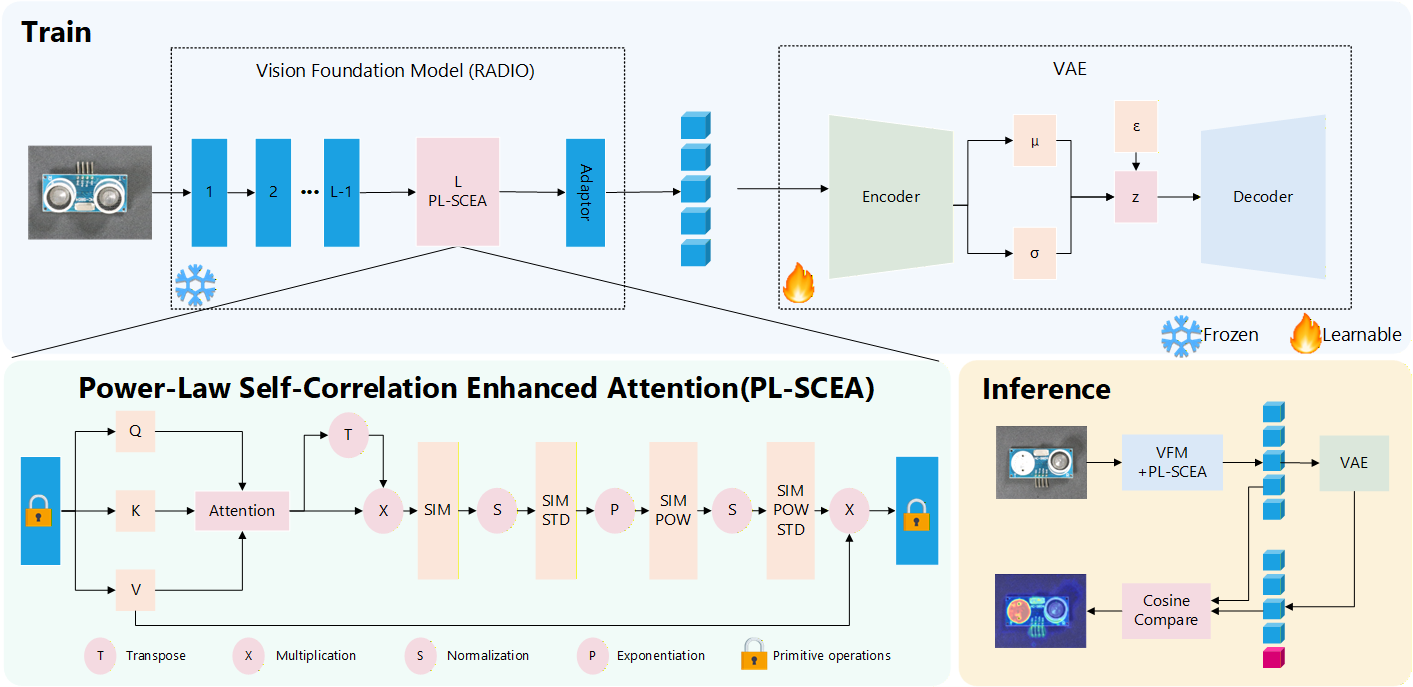}
    \caption{Overview of the proposed framework. The first stage reconfigures the final-layer attention computation of a frozen vision foundation model. PL-SCEA preserves QK semantic conditioning, constructs self-correlations over contextualized value features, and uses the reweighted relations to produce relation-aware patch representations. The second stage models these representations with a lightweight VAE. During inference, cosine discrepancies between the PL-SCEA-enhanced features and their deterministic VAE reconstructions are converted into an anomaly map. The two stages respectively control how relational deviations are represented and how deviations from learned normality are scored.}    
    \label{fig:complete_pipeline}
\end{figure*}

We evaluate the complete framework on MVTec AD and VisA across 1--8-shot settings. It maintains competitive image-level detection performance and improves pixel-level F1-max on MVTec AD by 4.1--4.8 percentage points over AnomalyDINO-S. Controlled comparisons further show that replacing QK attention with PL-SCEA improves localization with either the VAE or a memory bank in the MVTec AD 1-shot setting. Meanwhile, the PL-SCEA--VAE combination obtains the strongest pixel-level results in the component comparison. These findings suggest that attention reconfiguration and reconstruction-based normality modeling provide complementary benefits rather than attributing the complete improvement to either component alone. Figure~\ref{fig:complete_pipeline} summarizes the training and inference pipeline.
\textbf{Our contributions are summarized as follows:}
\begin{itemize}
\item We formulate and empirically examine the hypothesis that the attention computation of a frozen vision foundation model can be reconfigured to improve few-shot industrial anomaly localization, rather than treating pretrained token interactions as fixed.
\item We instantiate this design principle with PL-SCEA, which retains pretrained QK semantic conditioning while constructing token-adaptive, power-law-reweighted self-correlations without introducing additional trainable attention projections.
\item We combine relation-aware feature construction with reconstruction-based latent normality modeling. Experiments on MVTec AD and VisA show strong localization performance across 1--8-shot settings, while controlled comparisons with both a VAE and a memory bank indicate that the effect of PL-SCEA is not restricted to a single normality model.
\end{itemize}

\section{Related Work}

\noindent\textbf{Vision Foundation Models} transfer broad visual priors to downstream tasks. For anomaly detection, CLIP~\cite{clip} contributes vision-language alignment, DINO~\cite{dino} contributes self-supervised features, and SAM~\cite{sam} contributes dense region awareness. RADIO~\cite{radio} distills several teachers rather than relying on one pretraining objective. Its encoder combines the global semantics, local correspondence, and dense spatial cues of DINO, CLIP, and SAM. RADIOv3~\cite{PHI-S} further sharpens high-resolution dense features. We use a frozen RADIOv3 encoder and modify only how its existing tokens interact.

\noindent\textbf{Few-Shot Anomaly Detection}
Given a few normal reference images, few-shot industrial anomaly detection aims to learn category-specific normality and localize regions that deviate from it. We focus on low-level visual anomalies, such as scratches, cracks, contamination, and local structural damage, rather than semantic anomalies such as incorrect object categories. Because limited normal samples cannot cover variations in texture, pose, and structure, the central challenge is to construct a robust category-specific normality model from pretrained visual knowledge.

Cross-category methods transfer anomaly-detection knowledge to new categories through unsupervised meta-learning~\cite{MetaFormer}, hierarchical generation~\cite{HTDG}, or registration-based alignment~\cite{RegAD}. However, they generally require auxiliary category data or target-category adaptation.

Feature-matching methods encode normal images into patch features and use the nearest-neighbor distance to normal features as the anomaly score. PatchCore~\cite{patchcore} compresses the feature bank through coreset sampling, GraphCore~\cite{graphcore} models graph relations among support features, and AnomalyDINO~\cite{Anomalydino} shows that high-quality DINO features with nearest-neighbor retrieval can achieve strong few-shot performance. FastRecon~\cite{Fastrecon} reconstructs query features from support features and scores their reconstruction discrepancy. These methods typically remain dependent on reference-feature storage or retrieval.

Despite their different scoring mechanisms, these approaches primarily operate on patch representations that have already been formed by the pretrained backbone. They therefore address how normal features are stored, matched, or reconstructed, rather than whether the backbone's token interaction rule is itself appropriate for anomaly localization.

Vision--language methods compare image or patch features with normal and anomalous textual prompts. WinCLIP~\cite{WinCLIP}, PromptAD~\cite{PromptAD}, and InCTR~\cite{InCTRL} introduce multi-scale windows, learnable prompts, and normal-image context, respectively, while AnoPLe~\cite{AnoPLe} and KAG-prompt~\cite{KAG-prompt} further model interactions among visual prompts, textual prompts, and multi-level features. AnomalyGPT~\cite{AnomalyGPT} extends this paradigm to instruction-guided anomaly analysis. Although these methods can use language semantics to alleviate normal-data scarcity, their localization performance still depends on the quality of dense visual features.

\noindent\textbf{Attention Reshaping in Vision Transformers.} Final-layer attention has also been modified to improve local features for dense prediction. SCLIP~\cite{SCLIP} builds attention from query-query and key-key similarities, NACLIP~\cite{NACLIP} imposes a spatial-neighborhood prior, and ClearCLIP~\cite{ClearCLIP} combines same-type attention with a simplified architecture to limit interference from global semantics. For anomaly detection, AnomalyCLIP~\cite{AnomalyCLIP} substitutes value-value attention for the original QK relation to sharpen anomalous regions in CLIP features. These studies show that pretrained attention need not remain fixed when dense local representations are required. However, directly replacing QK attention with a same-type relation does not explain how pretrained semantic conditioning and anomaly-sensitive relational structure should be combined. We investigate this issue in few-shot industrial anomaly detection by retaining QK conditioning while reconfiguring the self-correlation structure of contextualized tokens.

\section{Attention Reconfiguration and Latent Normality Modeling}
Our framework separates few-shot anomaly detection into two complementary stages. The first stage concerns representation formation: PL-SCEA reconfigures the final-layer token interactions of a frozen VFM to produce relation-aware patch features. The second stage concerns normality estimation: Latent Normality Modeling reconstructs these features and uses their directional reconstruction discrepancies as anomaly evidence. This decomposition allows us to examine whether attention reconfiguration contributes beyond a particular downstream normality model.

\subsection{Token Correlations for Anomaly Detection}
\begin{figure}[t]
	\centering		
    \includegraphics[width=1\linewidth]{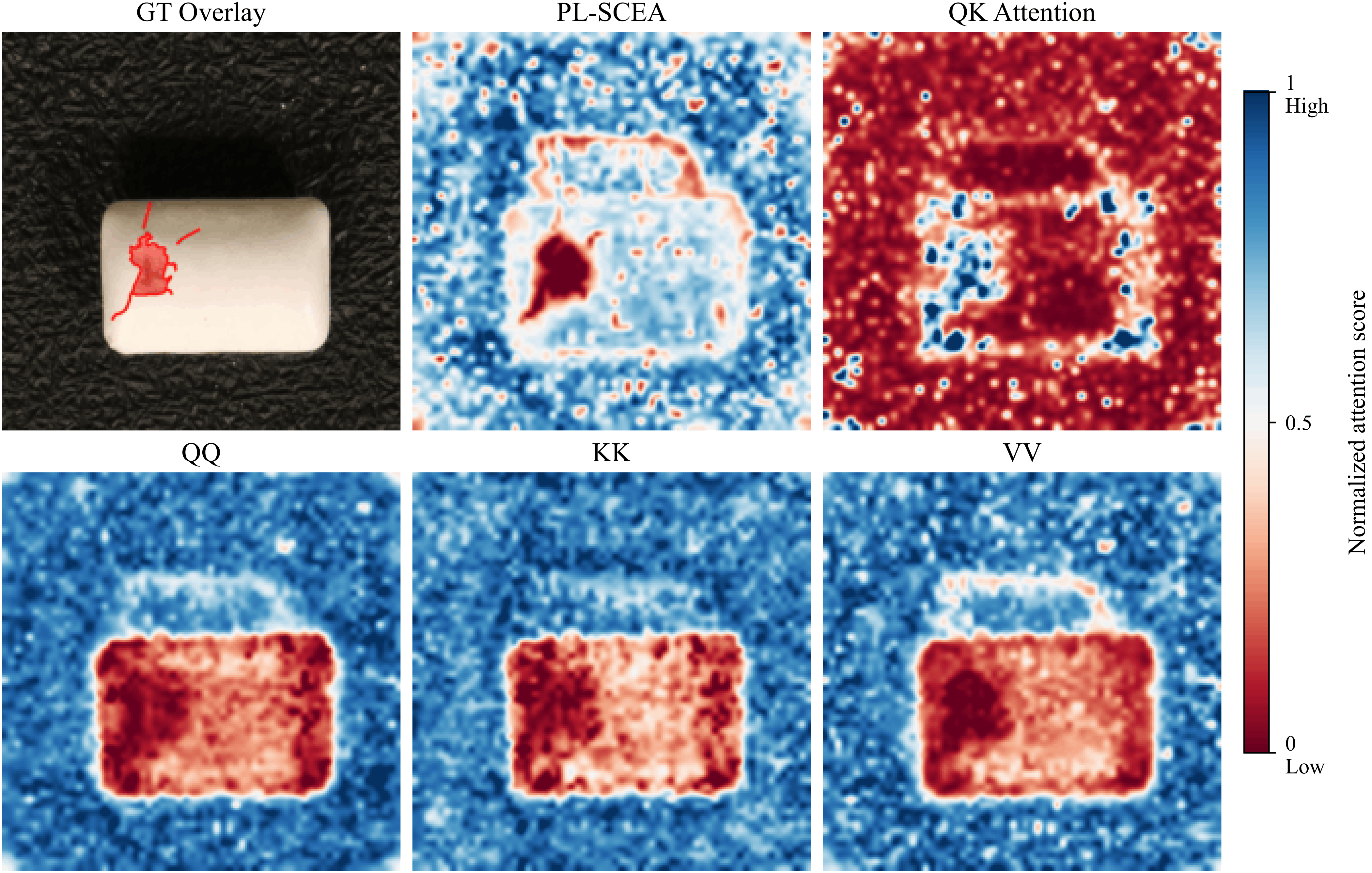}
    \caption{Comparison of anomaly responses produced by different attention branches. The ring-shaped responses in the non-PL-SCEA maps appear to be associated with the shadow above the object, which forms a strong intensity boundary.}
	\label{chewinggum_attention_comparison}
\end{figure} 

The final self-attention block directly influences how output patch representations aggregate information from the remaining tokens. Because the pretrained model is optimized for general-purpose representation learning, its original attention computation may emphasize semantic or high-contrast structures that are useful for recognition but less aligned with fine-grained anomaly localization.

Figure~\ref{chewinggum_attention_comparison} illustrates this mismatch in one example. The original QK attention retains semantic context but produces relatively fragmented responses, whereas QQ, KK, and VV correlations form more continuous relation maps. However, direct same-type correlations can also merge locally distinct patterns. The comparison indicates that changing the attention relation is not sufficient by itself; the form of the reconfiguration matters.

Tokens from repetitive textures and regular structures often exhibit consistent relational patterns, while local defects may disturb these patterns. This motivates an attention design that retains QK semantic conditioning but recalibrates token relations according to their relative correlation background. PL-SCEA implements this design by computing self-correlations over semantically contextualized value features and selectively reweighting the resulting relations.
\subsection{Power-Law Self-Correlation Enhanced Attention}
To implement this task-aligned attention reconfiguration, PL-SCEA first obtains semantically conditioned value representations using the original QK attention, extracts the token self-correlation structure from these representations, and constructs new attention relations through token-adaptive standardization and power-law reweighting. Rather than indiscriminately increasing token similarity, PL-SCEA emphasizes above-average self-correlations on a per-token basis, thereby increasing the relative contrast between strongly and weakly related tokens.

Let $\mathbf X\in\mathbb R^{N\times C}$ be the spatial-token input to the final attention layer, where $N$ is the number of spatial patch tokens and $C$ is the feature dimension. Prefix tokens are excluded from the PL-SCEA correlation construction and downstream patch-level anomaly scoring. For a multi-head attention module with $H$ heads, each head has dimension $d=C/H$. Using the pretrained QKV projections, we obtain $\mathbf Q,\mathbf K,\mathbf V\in\mathbb R^{N\times d}$. For clarity, we formulate the computation for a single head. The original attention is
\begin{equation}
    \mathbf A^{qk}=\operatorname{Softmax}
    \left(\frac{\mathbf Q\mathbf K^\top}{\sqrt d}\right).
\end{equation}
The contextualized value representation is then
\begin{equation}
    \mathbf Y=\mathbf A^{qk}\mathbf V.
\end{equation}
Unlike standard self-attention, we do not use $\mathbf Y$ directly as the final attention output. Instead, it serves as a semantically conditioned intermediate representation for estimating relational consistency among tokens.

\noindent\textbf{Semantically conditioned self-correlation.} To reduce the influence of feature magnitude and emphasize directional consistency among token representations, we first apply $L_2$ normalization to each contextualized token along the feature dimension:
\begin{equation}
    \tilde{\mathbf y}_i
    =
    \frac{\mathbf y_i}{\|\mathbf y_i\|_2+\epsilon},
\end{equation}
where $\epsilon$ ensures numerical stability. Pairwise self-correlation is then
\begin{equation}
    S_{ij}=\tilde{\mathbf y}_i^\top\tilde{\mathbf y}_j,
\end{equation}
where $S_{ij}$ is the cosine similarity between tokens $i$ and $j$ after the initial semantic aggregation.

\noindent\textbf{Token-adaptive correlation standardization.} The correlation distributions may differ substantially across query tokens and attention heads. For token $i$, PL-SCEA computes the row-wise mean and standard deviation:
\begin{equation}
    \mu_i=\frac{1}{N}\sum_{j=1}^{N}S_{ij},
    \qquad
    \sigma_i=
    \sqrt{\frac{1}{N}\sum_{j=1}^{N}(S_{ij}-\mu_i)^2}.
\end{equation}
The standardized correlation is
\begin{equation}
    \hat S_{ij}=\frac{S_{ij}-\mu_i}{\sigma_i+\epsilon}.
\end{equation}

\noindent\textbf{Power-law correlation enhancement.} This operation converts absolute cosine similarity into relational salience relative to the correlation background of the current token. When $\hat S_{ij}>0$, the relation between tokens $i$ and $j$ is stronger than the average correlation of token $i$. Each token is therefore calibrated against its own relational distribution rather than a single global criterion. Weak or moderately strong token interactions may introduce relational noise and obscure subtle local anomalies. PL-SCEA consequently retains only above-average correlations and enhances their contrast with a power-law transformation:
\begin{equation}
    R_{ij}=
    \left[\max(0,\hat S_{ij})\right]^{\gamma},
    \qquad \gamma>0.
\end{equation}
The exponent $\gamma>0$ controls the concentration of the retained positive standardized correlations. When $0<\gamma<1$, the power transformation compresses the relative differences between stronger and weaker positive correlations. When $\gamma=1$, it preserves the magnitudes of the positive standardized correlations while removing non-positive entries. When $\gamma>1$, it increases their relative contrast, assigning greater normalized weight to the strongest relations. We use $\gamma=2$ by default.

\noindent\textbf{Value reaggregation.} We normalize the enhanced correlations row-wise to obtain the final PL-SCEA weights:
\begin{equation}
    D_i=\sum_{j=1}^{N}R_{ij},
    \qquad
    A^{\mathrm{PL\text{-}SCEA}}_{ij}
    =
    \begin{cases}
        \dfrac{R_{ij}}{D_i}, & D_i>0,\\[8pt]
        \delta_{ij}, & D_i=0,
    \end{cases}
\end{equation}
where $\delta_{ij}$ is the Kronecker delta. The fallback branch retains the token itself when a row contains no positively salient relation. Therefore,
\begin{equation}
    A^{\mathrm{PL\text{-}SCEA}}_{ij}\geq0,
    \qquad
    \sum_{j=1}^{N}A^{\mathrm{PL\text{-}SCEA}}_{ij}=1.
\end{equation}
Finally, PL-SCEA uses the new attention weights to reaggregate the original value tokens:
\begin{equation}
    \mathbf O=\mathbf A^{\mathrm{PL\text{-}SCEA}}\mathbf V.
\end{equation}
The head-wise outputs are concatenated and passed through the frozen pretrained output projection. The surrounding residual connection, LayerNorm placement, and feed-forward/MLP sublayer remain unchanged from the original RADIOv3 block. Only the resulting spatial-token features are passed to Latent Normality Modeling and patch-level anomaly scoring.

\begin{table*}[t]
\centering
\small
\begin{tabular}{@{}llcccccc@{}}
\toprule
\multirow{2}{*}{Shots} & \multirow{2}{*}{Method} & \multicolumn{3}{c}{Image-level} & \multicolumn{3}{c}{Pixel-level} \\
\cmidrule(lr){3-5}\cmidrule(l){6-8}
& & AUROC & AUPR & $F_1$-max & AUROC & $F_1$-max & PRO \\
\midrule
\multirow{6}{*}{1-shot} & PatchCore~\cite{patchcore} & 83.4$\pm$3.0 & 92.2$\pm$1.5 & 90.5$\pm$1.5 & 92.0$\pm$1.0 & 50.4$\pm$2.1 & 79.7$\pm$2.0 \\
 & APRIL-GAN~\cite{APRIL-GAN} & 92.0$\pm$0.3 & 95.8$\pm$0.2 & 92.4$\pm$0.2 & 95.1$\pm$0.1 & 54.2$\pm$0.0 & 90.6$\pm$0.2 \\
 & WinCLIP+~\cite{WinCLIP} & 93.1$\pm$2.0 & 96.5$\pm$0.9 & 93.7$\pm$1.1 & 95.2$\pm$0.5 & 55.9$\pm$2.7 & 87.1$\pm$1.2 \\
 & AdaptCLIP~\cite{AdaptCLIP} & 94.5$\pm$0.5 & 97.5$\pm$0.1 & 95.0$\pm$0.0 & 94.3$\pm$0.1 & 54.0$\pm$0.7 & -- \\
 & AnomalyDINO-S~\cite{Anomalydino} & 96.6$\pm$0.4 & \textbf{98.2$\pm$0.2} & 95.8$\pm$0.5 & 96.8$\pm$0.1 & 60.2$\pm$1.1 & 92.7$\pm$0.1 \\
 & \textbf{PL-SCEA (Proposed)} & \textbf{96.9$\pm$0.9} & \textbf{98.2$\pm$0.8} & \textbf{95.9$\pm$0.7} & \textbf{97.5$\pm$0.1} & \textbf{64.3$\pm$1.4} & \textbf{93.7$\pm$0.2} \\
\midrule
\multirow{6}{*}{2-shot} & PatchCore~\cite{patchcore} & 86.3$\pm$3.3 & 93.8$\pm$1.7 & 92.0$\pm$1.5 & 93.3$\pm$0.6 & 53.0$\pm$1.7 & 82.3$\pm$1.3 \\
 & APRIL-GAN~\cite{APRIL-GAN} & 92.4$\pm$0.3 & 96.0$\pm$0.2 & 92.6$\pm$0.1 & 95.5$\pm$0.0 & 55.9$\pm$0.5 & 91.3$\pm$0.1 \\
 & WinCLIP+~\cite{WinCLIP} & 94.4$\pm$1.3 & 97.0$\pm$0.7 & 94.4$\pm$0.8 & 96.0$\pm$0.3 & 58.4$\pm$1.7 & 88.4$\pm$0.9 \\
 & AdaptCLIP~\cite{AdaptCLIP} & 95.7$\pm$0.6 & 97.9$\pm$0.2 & 95.4$\pm$0.1 & 94.5$\pm$0.0 & 55.0$\pm$0.3 & -- \\
 & AnomalyDINO-S~\cite{Anomalydino} & 96.9$\pm$0.7 & 98.2$\pm$0.5 & 96.1$\pm$0.3 & 97.0$\pm$0.2 & 61.0$\pm$0.5 & 93.1$\pm$0.2 \\
 & \textbf{PL-SCEA (Proposed)} & \textbf{97.0$\pm$1.0} & \textbf{98.6$\pm$0.4} & \textbf{96.7$\pm$0.0} & \textbf{97.9$\pm$0.0} & \textbf{65.7$\pm$0.5} & \textbf{94.2$\pm$0.2} \\
\midrule
\multirow{6}{*}{4-shot} & PatchCore~\cite{patchcore} & 88.8$\pm$2.6 & 94.5$\pm$1.5 & 92.6$\pm$1.6 & 94.3$\pm$0.5 & 55.0$\pm$1.9 & 84.3$\pm$1.6 \\
 & APRIL-GAN~\cite{APRIL-GAN} & 92.8$\pm$0.2 & 96.3$\pm$0.1 & 92.8$\pm$0.1 & 95.9$\pm$0.0 & 56.9$\pm$0.1 & 91.8$\pm$0.1 \\
 & WinCLIP+~\cite{WinCLIP} & 95.2$\pm$1.3 & 97.3$\pm$0.6 & 94.7$\pm$0.8 & 96.2$\pm$0.3 & 59.5$\pm$1.8 & 89.0$\pm$0.8 \\
 & AdaptCLIP~\cite{AdaptCLIP} & 96.6$\pm$0.3 & 98.4$\pm$0.2 & 96.0$\pm$0.0 & 94.8$\pm$0.1 & 56.8$\pm$0.7 & -- \\
 & AnomalyDINO-S~\cite{Anomalydino} & \textbf{97.7$\pm$0.2} & 98.7$\pm$0.1 & 96.6$\pm$0.0 & 97.2$\pm$0.1 & 61.8$\pm$0.1 & 93.4$\pm$0.1 \\
 & \textbf{PL-SCEA (Proposed)} & 97.3$\pm$1.1 & \textbf{98.9$\pm$0.3} & \textbf{96.9$\pm$0.2} & \textbf{98.0$\pm$0.2} & \textbf{66.3$\pm$1.0} & \textbf{94.6$\pm$0.4} \\
\midrule
\multirow{4}{*}{8-shot} & APRIL-GAN~\cite{APRIL-GAN} & 93.1$\pm$0.2 & 96.4$\pm$0.2 & 93.1$\pm$0.2 & 96.2$\pm$0.1 & 57.7$\pm$0.2 & 92.4$\pm$0.2 \\
 & WinCLIP+ ~\cite{WinCLIP}& 94.6$\pm$0.1 & -- & -- & -- & -- & -- \\
 & AnomalyDINO-S~\cite{Anomalydino} & \textbf{98.2$\pm$0.2} & 98.7$\pm$0.1 & \textbf{97.4$\pm$0.2} & 97.4$\pm$0.1 & 62.3$\pm$0.1 & 93.8$\pm$0.1 \\
 & \textbf{PL-SCEA (Proposed)} & 98.0$\pm$0.2 & \textbf{99.1$\pm$0.1} & \textbf{97.4$\pm$0.2} & \textbf{98.2$\pm$0.1} & \textbf{67.1$\pm$0.7} & \textbf{95.0$\pm$0.2} \\
\bottomrule
\end{tabular}
\caption{Few-shot anomaly-detection results on MVTec AD. Higher is better for all metrics. Values are mean $\pm$ standard deviation; -- indicates an unreported result. The highest reported mean within each shot is in \textbf{bold}.}
\label{tab:mvtec-results}
\end{table*}
\begin{figure*}[!t]
	\centering		
    \includegraphics[width=1\linewidth]{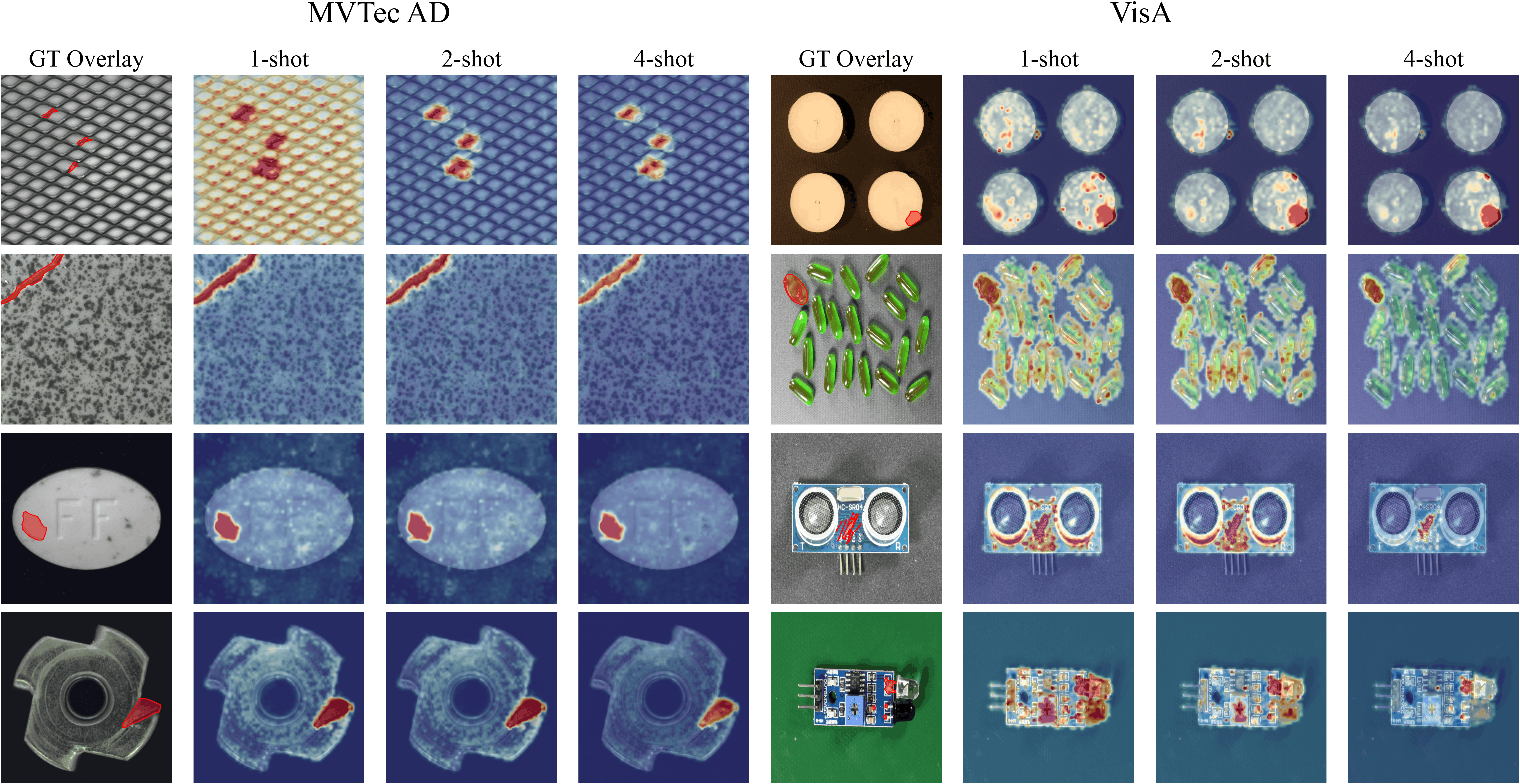}
    \caption{Qualitative results on the MVTec AD and VisA datasets.}
	\label{mvtec_visa_horizontal}
\end{figure*} 
\begin{table*}[!t]
\centering
\small
\begin{tabular}{@{}llcccccc@{}}
\toprule
\multirow{2}{*}{Shots} & \multirow{2}{*}{Method} & \multicolumn{3}{c}{Classification} & \multicolumn{3}{c}{Segmentation} \\
\cmidrule(lr){3-5}\cmidrule(l){6-8}
& & AUROC & AUPR & $F_1$-max & AUROC & $F_1$-max & PRO \\
\midrule
\multirow{6}{*}{1-shot} & PatchCore~\cite{patchcore} & 79.9$\pm$2.9 & 82.8$\pm$2.3 & 81.7$\pm$1.6 & 95.4$\pm$0.6 & 38.0$\pm$1.9 & 80.5$\pm$2.5 \\
 & WinCLIP+~\cite{WinCLIP} & 83.8$\pm$4.0 & 85.1$\pm$4.0 & 83.1$\pm$1.7 & 96.4$\pm$0.4 & 41.3$\pm$2.3 & 85.1$\pm$2.1 \\
 & AnomalyDINO-S~\cite{Anomalydino} & 87.4$\pm$1.2 & 89.0$\pm$1.0 & 84.3$\pm$0.5 & 97.8$\pm$0.1 & 45.1$\pm$0.9 & 92.5$\pm$0.5 \\
 & AdaptCLIP~\cite{AdaptCLIP} & 90.5$\pm$1.2 & 92.3$\pm$0.9 & 86.5$\pm$1.0 & 96.8$\pm$0.0 & 44.6$\pm$0.4 & -- \\
 & APRIL-GAN~\cite{APRIL-GAN} & \textbf{91.2$\pm$0.8} & \textbf{93.3$\pm$0.8} & 86.9$\pm$0.6 & 96.0$\pm$0.0 & 38.5$\pm$0.3 & 90.0$\pm$0.1 \\
 & \textbf{PL-SCEA (Proposed)} & 91.1$\pm$2.4 & 92.3$\pm$1.6 & \textbf{88.3$\pm$1.4} & \textbf{98.0$\pm$0.1} & \textbf{46.8$\pm$0.8} & \textbf{93.1$\pm$0.2} \\
\midrule
\multirow{6}{*}{2-shot} & PatchCore~\cite{patchcore} & 81.6$\pm$4.0 & 84.8$\pm$3.2 & 82.5$\pm$1.8 & 96.1$\pm$0.5 & 41.0$\pm$3.9 & 82.6$\pm$2.3 \\
 & WinCLIP+~\cite{WinCLIP} & 84.6$\pm$2.4 & 85.8$\pm$2.7 & 83.0$\pm$1.4 & 96.8$\pm$0.3 & 43.5$\pm$3.3 & 86.2$\pm$1.4 \\
 & AnomalyDINO-S ~\cite{Anomalydino}& 89.7$\pm$1.3 & 90.7$\pm$0.8 & 86.3$\pm$1.2 & 98.0$\pm$0.1 & 47.6$\pm$0.5 & 93.4$\pm$0.6 \\
 & APRIL-GAN~\cite{APRIL-GAN} & 92.2$\pm$0.3 & 94.2$\pm$0.3 & 87.7$\pm$0.3 & 96.2$\pm$0.0 & 39.3$\pm$0.2 & 90.1$\pm$0.1 \\
 & AdaptCLIP~\cite{AdaptCLIP} & 92.2$\pm$0.8 & 93.6$\pm$0.6 & 88.0$\pm$0.7 & 97.1$\pm$0.0 & 46.1$\pm$0.4 & -- \\
 & \textbf{PL-SCEA (Proposed)} & \textbf{94.0$\pm$0.3} & \textbf{94.7$\pm$0.2} & \textbf{89.9$\pm$0.6} & \textbf{98.3$\pm$0.1} & \textbf{48.7$\pm$1.2} & \textbf{94.2$\pm$0.1} \\
\midrule
\multirow{6}{*}{4-shot} & PatchCore~\cite{patchcore} & 85.3$\pm$2.1 & 87.5$\pm$2.1 & 84.3$\pm$1.3 & 96.8$\pm$0.3 & 43.9$\pm$3.1 & 84.9$\pm$1.4 \\
 & WinCLIP+~\cite{WinCLIP} & 87.3$\pm$1.8 & 88.8$\pm$1.8 & 84.2$\pm$1.6 & 97.2$\pm$0.2 & 47.0$\pm$3.0 & 87.6$\pm$0.9 \\
 & APRIL-GAN~\cite{APRIL-GAN} & 92.6$\pm$0.4 & 94.5$\pm$0.3 & 88.4$\pm$0.5 & 96.2$\pm$0.0 & 40.0$\pm$0.1 & 90.2$\pm$0.1 \\
 & AnomalyDINO-S~\cite{Anomalydino} & 92.6$\pm$0.9 & 92.9$\pm$0.7 & 88.8$\pm$0.9 & 98.2$\pm$0.0 & 49.4$\pm$0.3 & 94.1$\pm$0.1 \\
 & AdaptCLIP~\cite{AdaptCLIP} & 93.1$\pm$0.2 & 94.3$\pm$0.2 & 88.5$\pm$0.2 & 97.3$\pm$0.0 & 47.2$\pm$0.5 & -- \\
 & \textbf{PL-SCEA (Proposed)} & \textbf{94.9$\pm$0.3} & \textbf{95.5$\pm$0.2} & \textbf{91.1$\pm$0.5} & \textbf{98.6$\pm$0.0} & \textbf{50.6$\pm$1.0} & \textbf{95.1$\pm$0.1} \\
\midrule
\multirow{4}{*}{8-shot} & WinCLIP+~\cite{WinCLIP} & 85.0$\pm$0.0 & -- & -- & -- & -- & -- \\
 & APRIL-GAN~\cite{APRIL-GAN} & 93.0$\pm$0.2 & 94.9$\pm$0.3 & 88.8$\pm$0.2 & 96.3$\pm$0.0 & 40.2$\pm$0.1 & 90.2$\pm$0.0 \\
 & AnomalyDINO-S~\cite{Anomalydino} & 93.8$\pm$0.3 & 94.3$\pm$0.4 & 90.0$\pm$0.1 & 98.4$\pm$0.0 & 51.1$\pm$0.4 & 94.8$\pm$0.2 \\
 & \textbf{PL-SCEA (Proposed)} & \textbf{96.5$\pm$0.2} & \textbf{97.0$\pm$0.2} & \textbf{92.9$\pm$0.2} & \textbf{98.9$\pm$0.0} & \textbf{52.1$\pm$0.3} & \textbf{95.6$\pm$0.0} \\
\bottomrule
\end{tabular}
\caption{Few-shot anomaly-detection results on VisA. Higher is better for all metrics. Values are mean $\pm$ standard deviation; -- indicates an unreported result. The highest reported mean within each shot is in \textbf{bold}.}
\label{tab:visa-results}
\end{table*}
\subsection{Reconstruction-Based Latent Normality Modeling}
After attention reconfiguration, Latent Normality Modeling (LNM) uses a lightweight VAE~\cite{VAE} to model the PL-SCEA-enhanced patch representations. Its role is distinct from that of PL-SCEA: the attention module determines how relational evidence is represented, whereas LNM learns a compact category-specific model of normality and converts reconstruction discrepancies into anomaly scores. Let $\mathbf{x}_i$ denote the $i$-th enhanced patch feature. During training, $\mathbf{x}_i$ is first $L_2$-normalized to obtain the clean reconstruction target $\bar{\mathbf{x}}_i$. Gaussian noise is then added, followed by another normalization, to obtain the encoder input $\widetilde{\mathbf{x}}_i$.

The encoder $E_\phi$ maps $\widetilde{\mathbf{x}}_i$ to the mean $\boldsymbol{\mu}_i$ and standard deviation $\boldsymbol{\sigma}_i$ of the Gaussian posterior $q_\phi(\mathbf{z}_i\mid\widetilde{\mathbf{x}}_i)$. During training, the latent variable $\mathbf{z}_i$ is sampled using reparameterization, and the decoder $D_\theta$ reconstructs the normalized patch feature $\widehat{\mathbf{x}}_i$. The LNM objective combines cosine reconstruction loss with KL regularization:

\begin{equation}
   \begin{aligned}
    \mathcal L_{\mathrm{LNM}}
    &=
    \underbrace{
    \frac{1}{B}
    \sum_{i=1}^{B}
    \left[
    1-
    \operatorname{sim}_{\mathrm{cos}}
    \left(
    \widehat{\mathbf x}_i,
    \bar{\mathbf x}_i
    \right)
    \right]
    }_{\mathcal L_{\mathrm{cos}}}
    \\
    &\quad+
    \beta
    \underbrace{
    \frac{1}{B}
    \sum_{i=1}^{B}
    D_{\mathrm{KL}}
    \left[
    q_\phi
    \left(
    \mathbf z_i\mid\widetilde{\mathbf x}_i
    \right)
    \|
    \mathcal N(\mathbf 0,\mathbf I)
    \right]
    }_{\mathcal L_{\mathrm{KL}}}.
    \end{aligned}
\end{equation}

The reconstruction term encourages the model to preserve the directional structure of normal features, while the KL term regularizes the latent distribution toward a standard Gaussian. We use a small $\beta$ to limit the risk that excessive regularization suppresses fine-grained variations among normal patches.

At inference, no stochastic sampling is performed. The encoder receives the normalized test feature $\bar{\mathbf{x}}_i$, and its posterior mean $\boldsymbol{\mu}_i$ is passed to the decoder for deterministic reconstruction. The raw and standardized anomaly scores are

\begin{equation}
    a_i
    =
    1-
    \operatorname{sim}_{\mathrm{cos}}
    \left(
    \bar{\mathbf{x}}_i,
    \operatorname{Norm}_2
    \left[
    D_\theta(\boldsymbol{\mu}_i)
    \right]
    \right),
\end{equation}
\begin{equation}
    \tilde{a}_i
    =
    \frac{a_i-m_a}{s_a+\epsilon},
\end{equation}

where $m_a$ and $s_a$ are the mean and standard deviation of the reconstruction scores computed from normal training patches. Under this reconstruction-based modeling assumption, normal features are expected to produce smaller directional discrepancies than anomalous features. We therefore use the standardized reconstruction discrepancy as the patch-level anomaly score.

Considering only the category-specific normality state, LNM compresses a variable number of normal patch features into the fixed-size parameter set $\Theta=\{\phi,\theta\}$, where $\phi$ and $\theta$ denote the encoder and decoder parameters, respectively. Its additional storage complexity is therefore $O(|\Theta|)$ and does not grow with the number of reference samples. In contrast, the corresponding memory-bank state grows with the number of stored patch features. The frozen backbone, which is shared by both alternatives, is excluded from this comparison.

\section{Experiments}

\noindent\textbf{Datasets.}
MVTec AD~\cite{mvtec} contains over 5,000 high-resolution images from 15 texture and object categories, with image-level anomaly labels and pixel-level masks. VisA~\cite{visa} contains 9,621 normal and 1,200 anomalous images from products with complex structures and positional variation, covering scratches, dents, discoloration, cracks, and structural defects.

\noindent\textbf{Evaluation Metrics.}
We report image-level AUROC, AUPR, and F1-max, and pixel-level AUROC, F1-max, and PRO. PRO measures region overlap across thresholds. All metrics are percentages. For PL-SCEA, the main results in Tables~\ref{tab:mvtec-results} and~\ref{tab:visa-results} are category macro-averages over three support-set seeds, reported as mean $\pm$ sample standard deviation. Baseline entries are taken from the cited sources; because support-set selection and implementation details may differ, comparisons with our local results are descriptive rather than strictly controlled.

\noindent\textbf{Implementation Details.}
We follow AnomalyDINO~\cite{Anomalydino} for preprocessing. A frozen RADIOv3-B serves as the encoder, with PL-SCEA ($\gamma=2$) replacing self-attention in its final Transformer block. $L_2$-normalized patch features are modeled by a lightweight VAE with a 64-dimensional latent space. We train for up to 600 epochs using AdamW with a learning rate of $8\times10^{-4}$, weight decay of $1\times10^{-5}$, and batch size of 512 on one 12-GB NVIDIA RTX 3080 Ti. At inference, patch scores are standardized using normal training reconstruction errors and Gaussian-smoothed; the image score is the mean of the top 1\% patch scores.

\subsection{Quantitative Results}

\noindent\textbf{MVTec AD.}
Among the methods listed in Table~\ref{tab:mvtec-results}, PL-SCEA obtains the highest or tied-highest reported mean on all image- and pixel-level metrics in the 1- and 2-shot settings. With 4 and 8 shots, its image-level AUROC is respectively 0.4 and 0.2 points below that of AnomalyDINO-S, while it retains the highest reported means on all pixel-level metrics and the highest or tied-highest image-level AUPR and F1-max. In particular, PL-SCEA improves pixel-level F1-max over AnomalyDINO-S by 4.1--4.8 points across the four shot settings. These results indicate that the complete framework provides consistent localization gains while maintaining competitive image-level detection.

\noindent\textbf{VisA.}
Among the methods listed in Table~\ref{tab:visa-results}, the complete framework obtains the highest reported means on all classification and segmentation metrics in the 2-, 4-, and 8-shot settings. With one shot, it obtains the highest classification F1-max and all three pixel-level metrics, while its image-level AUROC and AUPR are respectively 0.1 and 1.0 points below the highest reported means. The reported performance of PL-SCEA increases with the number of normal references across all six metrics.

\subsection{Qualitative Analysis}

Figure~\ref{mvtec_visa_horizontal} presents qualitative examples covering fine scratches, local damage, and larger structural defects on MVTec AD and VisA. In the displayed cases, the predicted responses overlap the annotated anomalous regions across different defect shapes and scales. The 4-shot maps also appear more concentrated than their 1-shot counterparts, with fewer responses in normal textures and backgrounds. These observations are consistent with the quantitative localization results, although the qualitative examples alone do not establish performance across all categories.

\subsection{Ablation Studies}
Unless otherwise stated, all ablations use a fixed MVTec AD 1-shot support set and report single-run category macro-averages over 15 categories. These controlled comparisons are not numerically interchangeable with the three-seed means in Table~\ref{tab:mvtec-results}.
\textbf{Effect of Attention Reconfiguration.}
Table~\ref{tab:attention_ablation} examines whether changing the attention computation is sufficient by itself and whether the form of the reconfiguration matters. Among the tested formulations, PL-SCEA obtains the highest image-level AUROC, image-level AUPR, pixel-level AUROC, and pixel-level F1-max. Relative to the original QK attention, it improves pixel-level F1-max by 3.78 points. Direct QQ, KK, and VV replacements do not reproduce the same overall gains, and VV reduces pixel-level F1-max by 6.35 points relative to QK. These results show that attention modification is not universally beneficial; retaining QK conditioning while selectively reweighting contextualized self-correlations is important under this setting. QQ obtains the highest PRO of 93.77\%, while PL-SCEA remains within 0.03 points at 93.74\%.

\begin{table}[!htbp]
\centering
\small
\setlength{\tabcolsep}{3.2pt}
\begin{tabular}{lccccc}
\toprule
Attn. & I-AUROC & I-AUPR & P-AUROC & P-F1 & PRO \\
\midrule
QK & 96.54 & 98.06 & 97.12 & 61.54 & 93.55 \\
QQ & 96.65 & 98.31 & 97.14 & 63.81 & \textbf{93.77} \\
KK & 95.82 & 97.30 & 97.14 & 61.74 & 93.37 \\
VV & 92.44 & 95.34 & 95.45 & 55.19 & 91.53 \\
PL-SCEA & \textbf{97.26} & \textbf{98.65} & \textbf{97.56} & \textbf{65.32} & 93.74 \\
\bottomrule
\end{tabular}
\caption{Ablation of attention variants on MVTec AD under the fixed single-seed 1-shot setting. Values are category macro-averages over 15 categories.}
\label{tab:attention_ablation}
\end{table}

\begin{table}[!htbp]
\centering
\small
\setlength{\tabcolsep}{2.5pt}
\begin{tabular}{lcccc}
\toprule
Layer & \shortstack{I-AUROC} & \shortstack{P-AUROC} & \shortstack{P-F1} & PRO \\
\midrule
2 & 96.0 & 97.1 & 62.5 & 93.4 \\
4 & 95.0 & 97.0 & 61.5 & 91.9 \\
6 & 95.6 & 97.1 & 62.3 & 92.9 \\
8 & 95.8 & 96.8 & 60.7 & 92.4 \\
10 & 97.0 & 97.3 & 63.3 & 93.5 \\
{\bfseries 12} & {\bfseries 97.3} & {\bfseries 97.6} & {\bfseries 65.3} & {\bfseries 93.7} \\
\bottomrule
\end{tabular}
\caption{Effect of the PL-SCEA insertion layer on MVTec AD under the fixed single-seed 1-shot setting. Values are rounded to one decimal.}
\label{tab:layer}
\end{table}

\textbf{Effect of the Insertion Layer.}
Table~\ref{tab:layer} shows that applying PL-SCEA to the final RADIOv3 layer obtains the highest reported values on all selected metrics, reaching 97.3\% image-level AUROC, 97.6\% pixel-level AUROC, 65.3\% pixel-level F1-max, and 93.7\% PRO. Compared with layer 2, the final layer improves image-level AUROC and pixel-level F1-max by 1.3 and 2.8 points, respectively. This empirical trend is consistent with the hypothesis that later, more contextualized features provide token relations better suited to the proposed reconfiguration. Based on these results, we apply PL-SCEA only to the final Transformer block.

\begin{table}[!htbp]
\centering
\small
\setlength{\tabcolsep}{1.2pt}
\begin{tabular}{llcccc}
\toprule
Attn. & Model & I-AUROC & P-AUROC & P-F1 & PRO \\
\midrule
PL-SCEA & VAE & 97.3 & {\bfseries 97.6} & {\bfseries 65.3} & {\bfseries 93.7} \\
QK & VAE & 96.5 & 97.1 & 61.5 & 93.5 \\
PL-SCEA & Memory bank & {\bfseries 97.5} & 97.0 & 64.9 & 93.1 \\
QK & Memory bank & 97.1 & 96.3 & 61.4 & 92.9 \\
\bottomrule
\end{tabular}
\caption{Complementarity between PL-SCEA and the normality model on MVTec AD under the fixed single-seed 1-shot setting. QK denotes the original attention. Values are rounded to one decimal; the VAE rows correspond to Table~\ref{tab:attention_ablation}.}
\label{tab:components}
\end{table}

\textbf{Complementarity with Normality Modeling.}
Table~\ref{tab:components} separates the effects of attention formulation and normality modeling. Relative to QK attention, PL-SCEA improves pixel-level F1-max by 3.8 points with the VAE and by 3.5 points with the memory bank. It also improves pixel-level AUROC and PRO under both normality models. The effect of attention reconfiguration is therefore observed with both tested downstream alternatives rather than being restricted to the VAE. Meanwhile, pairing PL-SCEA with the VAE obtains the highest pixel-level AUROC, F1-max, and PRO, whereas the memory-bank variant obtains a 0.2-point higher image-level AUROC. These results suggest complementary roles for relation-aware feature construction and reconstruction-based normality modeling, but do not establish a strictly synergistic interaction between them.

\begin{figure}[!t]
	\centering
    \includegraphics[width=0.9\linewidth]{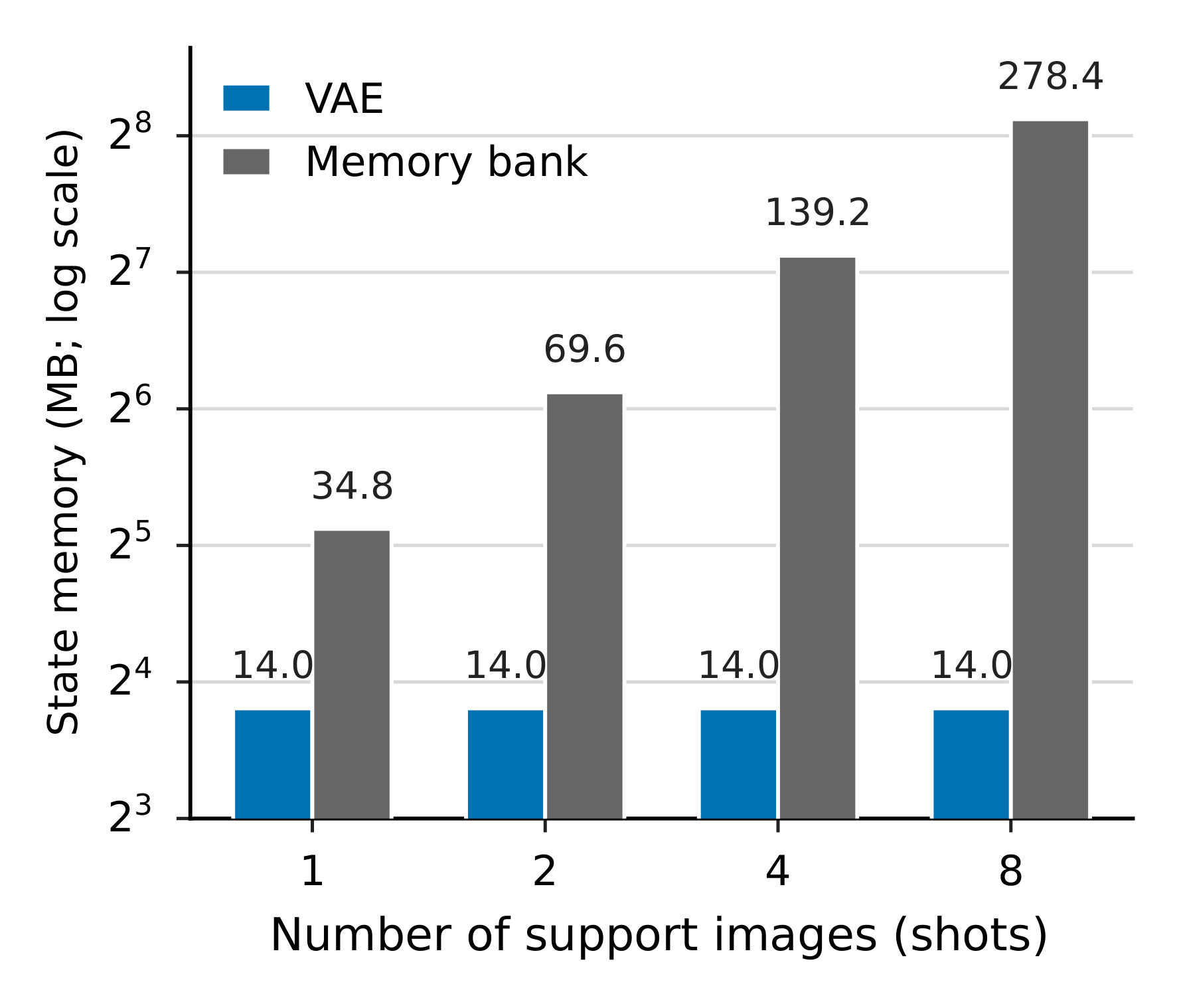}
    \caption{Category-specific inference-state memory across different few-shot settings on MVTec AD. The shared frozen backbone is excluded.}
	\label{fig_efficiency_memory_bar}

    \vspace{0.5em}
    \includegraphics[width=0.9\linewidth]{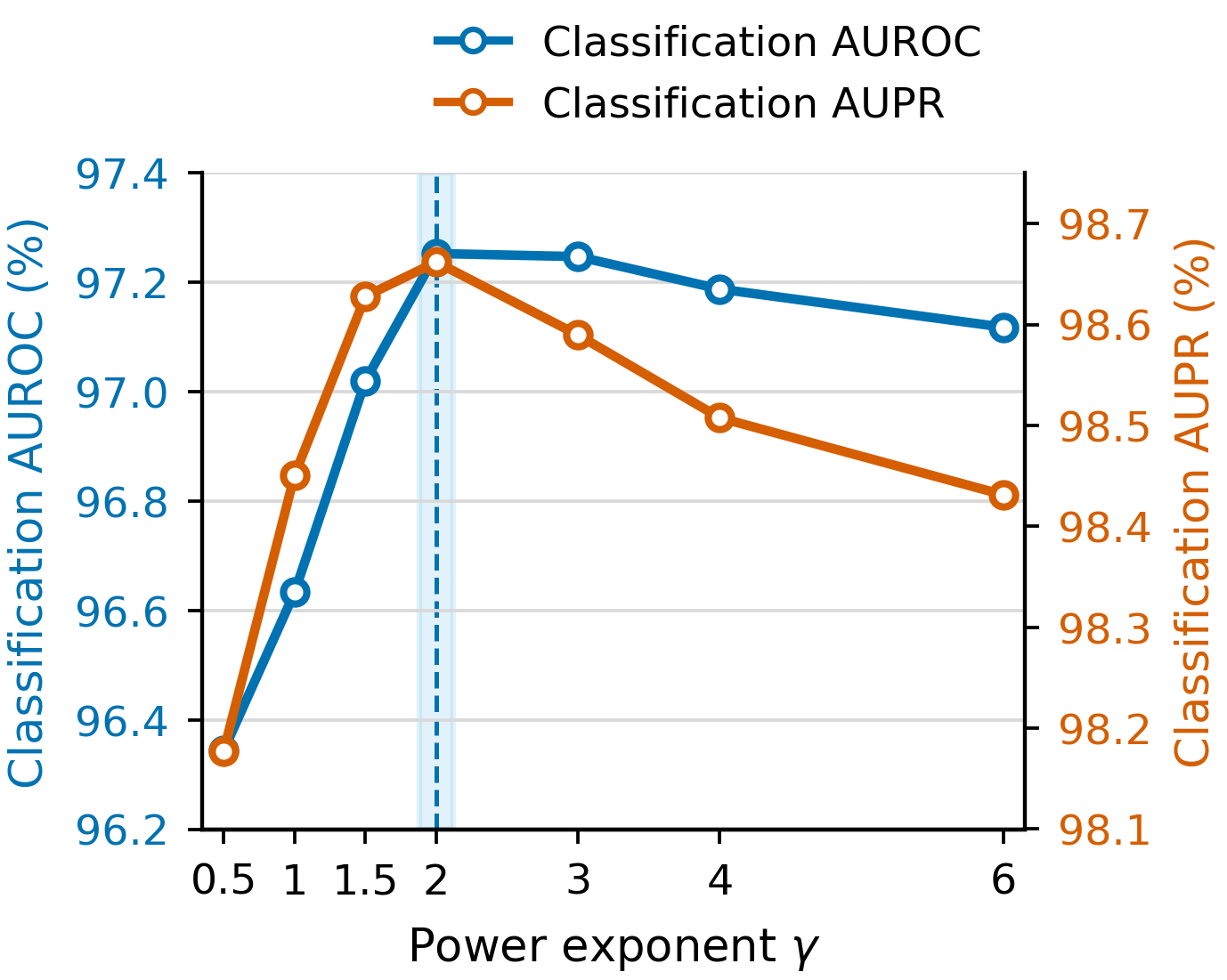}
    \caption{Effect of the power exponent $\gamma$ on image-level AUROC and AUPR on MVTec AD under the fixed single-seed 1-shot setting.}
	\label{fig_gamma_ablation_classification_ranking}
\end{figure}

\textbf{Inference-State Memory.}
Figure~\ref{fig_efficiency_memory_bar} compares the category-specific inference state required by the two normality models. The VAE maintains a fixed 14.0-MB footprint across 1--8 shots, whereas the memory bank grows linearly from 34.8 to 278.4 MB. At 8 shots, the VAE requires approximately 5.0\% of the memory used by the memory bank, corresponding to a 95.0\% reduction. This result highlights the storage benefit of reconstruction-based parametric normality modeling; it does not include the frozen backbone shared by both alternatives.

\textbf{Effect of the Power Exponent.}
Figure~\ref{fig_gamma_ablation_classification_ranking} shows that the reported image-level AUROC and AUPR increase as $\gamma$ changes from 0.5 to 2, reaching 97.25\% and 98.66\%, respectively. Both metrics are lower at larger tested values; at $\gamma=6$, they decrease to 97.12\% and 98.43\%. The results therefore favor a moderate exponent. This pattern suggests that moderate reweighting can emphasize informative relations, whereas stronger concentration may suppress complementary contextual information. We use $\gamma=2$ in the remaining experiments.

\section{Conclusion}
This study indicates that the attention computation of a frozen vision foundation model is not merely an inherited implementation detail; when reconfigured in a task-aligned manner, it can improve few-shot industrial anomaly localization. PL-SCEA operationalizes this principle by retaining pretrained QK conditioning while standardizing and power-law reweighting the self-correlations of contextualized tokens. Reconstruction-based latent normality modeling complements this representation stage by compactly summarizing normal features and converting their reconstruction discrepancies into anomaly scores. Results on MVTec AD and VisA support the effectiveness of the complete framework, while the MVTec AD 1-shot ablations show that the localization gains of PL-SCEA persist with either a VAE or a memory bank. Within the tested datasets and backbone, these findings support attention reconfiguration as a useful design direction for adapting pretrained representations to localized anomaly detection.


\bibliography{references}


\end{document}